%% file: acl_latex.tex
\pdfoutput=1

\documentclass[11pt]{article}

\usepackage[final]{acl}

\usepackage{times}
\usepackage{latexsym}
\usepackage[utf8]{inputenc}
\usepackage[mathletters]{ucs}
\usepackage{hyperref}
\usepackage{url}
\usepackage{graphicx}
\usepackage{tabularx} 
\usepackage{wrapfig}  
\usepackage{booktabs}
\usepackage{multirow}
\usepackage{xspace}
\usepackage{tablefootnote}
\usepackage{caption}
\usepackage{subcaption}
\usepackage{mathtools}
\usepackage{algorithm}
\usepackage{algpseudocode}
\usepackage{inconsolata}
\usepackage{microtype} 
\usepackage{euflag}
\usepackage{wrapfig} %
\usepackage{pifont}%
\usepackage{color,soul}
\usepackage{xcolor}
\usepackage{dirtytalk}
\usepackage[T5,T3,T1]{fontenc}
\usepackage[hang,flushmargin]{footmisc}
\usepackage{comment}

\usepackage{siunitx}
\usepackage{colortbl}

\colorlet{tableheadcolor}{gray!100} %
\colorlet{tablerowcolor}{gray!20} %
\newcommand{\tc}{\textcolor{tableheadcolor}} %

\input{latex/custom/custom_packages}
\input{latex/custom/custom_commands}

\title{Beyond Text Compression: Evaluating Tokenizers Across Scales}

\author{
    Jonas F. Lotz\thanks{Work done during an internship at Apple.} \\
    University of Copenhagen, Denmark \& \\
    ROCKWOOL Foundation Research Unit\\
    \texttt{jonasf.lotz@di.ku.dk} \\
    \And
    António V. Lopes \quad Stephan Peitz \\
    \textbf{Hendra Setiawan} \quad \textbf{Leonardo Emili} \\
    Apple \\
    \texttt{\{avlopes, speitz, hendra, lemili\}} \\
    \texttt{@apple.com}
}

\begin{document}
\maketitle

\input{latex/sections/0_abstract}

\input{latex/sections/1_intro}

\input{latex/sections/2_tokenizer}

\input{latex/sections/3_approach}

\input{latex/sections/4_results}

\input{latex/sections/5_predicting}

\input{latex/sections/6_discussion}

\input{latex/sections/7_related}

\input{latex/sections/8_conclusion}

\input{latex/sections/9_limitations}

\input{latex/sections/10_acknow}

\input{acl_latex.bbl}
\newpage
\appendix
\input{latex/sections/11_appendix}

\end{document}

%% file: latex/custom/custom_packages.tex
\usepackage{xcolor}
\usepackage{lipsum}
\usepackage{booktabs}
\usepackage{multicol}
\usepackage{multirow}
\usepackage{graphicx}
\usepackage{tabularx}
\usepackage{float}
\usepackage{bm}
\usepackage{todonotes}
\usepackage{xargs}
\usepackage{xspace}
\usepackage{amsmath}
\usepackage{cleveref}
\usepackage{subcaption}
\usepackage{mathtools}
\usepackage{bigstrut}
\usepackage{tikz}
\usetikzlibrary{shapes}
\usepackage{siunitx}
\usepackage{amssymb}
\usepackage{arydshln}
\usepackage{listings}
\usepackage{xfrac}

\usepackage{caption}
\usepackage{dcolumn}
\newcolumntype{L}{D{.}{.}{2,1}}

\setlipsum{%
  par-before = \begingroup\color{gray},
  par-after = \endgroup
}
\usepackage[inline]{enumitem}
\usepackage{tabularx}

%% file: latex/custom/custom_commands.tex
\makeatletter\newcommand*{\etc}{%
	\@ifnextchar{.}%
	{etc}%
	{etc.\@\xspace}%
}
\makeatother

\newcommand{\lde}{\textsc{de}\@\xspace}
\newcommand{\len}{\textsc{en}\@\xspace}
\newcommand{\lcs}{\textsc{cs}\@\xspace}
\newcommand{\lru}{\textsc{ru}\@\xspace}
\newcommand{\lzh}{\textsc{zh}\@\xspace}

\newcommand{\falcon}{\textsc{Falcon}\@\xspace}
\newcommand{\gpt}{\textsc{GPT-2}\@\xspace}
\newcommand{\gptneo}{\textsc{GPT-NeoX}\@\xspace}
\newcommand{\llama}{\textsc{Phi-3-mini}\@\xspace}
\newcommand{\aya}{\textsc{Aya 23} \@\xspace}

\newcommand{\phthreesmall}{\textsc{tiktoken}\@\xspace}

\newcommand{\xsum}{\textsc{X-Sum}\@\xspace}

\definecolor{red}{RGB}{141,45,57}
\definecolor{dark}{RGB}{55,65,74}
\definecolor{blue}{RGB}{0,105,170}
\definecolor{gold}{RGB}{174,159,109}
\definecolor{gray}{RGB}{175,179,183}
\definecolor{darkgreen}{RGB}{50,110,30}

\definecolor{ptgreen}{RGB}{213,232,212}

\makeatletter
\newcommand*\standardbin{+}
\newcommand*\tabularbin[1]{%
  \mathbin{\mathpalette{\@tabularsym\standardbin}{#1}}%
}
\newcommand*\@tabularsym[3]{%
  \setbox\z@\hbox{$#2#1\m@th$}%
  \hbox to\wd\z@{\hss$#2#3\m@th$\hss}%
}
\makeatother

\newcolumntype{Y}{>{\centering\arraybackslash}X}

\newcommandx{\hendra}[2][1=]{\vspace{0.2cm} \todo[linecolor=blue,backgroundcolor=blue!15,bordercolor=blue, #1]{\textbf{Hendra:} #2}}
\newcommandx{\jonas}[2][1=]{\vspace{0.2cm} \todo[linecolor=gold,backgroundcolor=gold!25,bordercolor=gold, #1]{\textbf{Jonas:} #2}}
\newcommandx{\antonio}[2][1=]{\vspace{0.2cm} \todo[linecolor=darkgreen,backgroundcolor=darkgreen!25,bordercolor=darkgreen, #1]{\textbf{Antonio:} #2}}
\newcommandx{\stephan}[2][1=]{\vspace{0.2cm} \todo[linecolor=yellow,backgroundcolor=yellow!25,bordercolor=yellow, #1]{\textbf{Stephan:} #2}}
\newcommandx{\leonardo}[2][1=]{\vspace{0.2cm} \todo[linecolor=brown,backgroundcolor=brown!25,bordercolor=brown, #1]{\textbf{Leonardo:} #2}}

\newcommand{\todotemplate}{\textcolor{red}{TODO}\xspace}
\newcommandx{\thendra}[2][1=]{\vspace{0.2cm} \todo[linecolor=blue,backgroundcolor=blue!15,bordercolor=blue, #1]{\textbf{\todotemplate @Hendra:} #2}}
\newcommandx{\tjonas}[2][1=]{\vspace{0.2cm} \todo[linecolor=gold,backgroundcolor=gold!25,bordercolor=gold, #1]{\textbf{\todotemplate @Jonas:} #2}}
\newcommandx{\tantonio}[2][1=]{\vspace{0.2cm} \todo[linecolor=darkgreen,backgroundcolor=darkgreen!25,bordercolor=darkgreen, #1]{\textbf{\todotemplate @Antonio:} #2}}
\newcommandx{\tstephan}[2][1=]{\vspace{0.2cm} \todo[linecolor=yellow,backgroundcolor=yellow!25,bordercolor=yellow, #1]{\textbf{\todotemplate @Stephan:} #2}}
\newcommandx{\tleonardo}[2][1=]{\vspace{0.2cm} \todo[linecolor=brown,backgroundcolor=brown!25,bordercolor=brown, #1]{\textbf{\todotemplate @Brown:} #2}}

\newcommandx{\antonioinline}[2][1=]{\vspace{0.2cm} \todo[linecolor=darkgreen,backgroundcolor=darkgreen!25,bordercolor=darkgreen, inline, #1]{\textbf{Antonio:} #2}}

\makeatletter
\def\adl@drawiv#1#2#3{%
        \hskip.5\tabcolsep
        \xleaders#3{#2.5\@tempdimb #1{1}#2.5\@tempdimb}%
                #2\z@ plus1fil minus1fil\relax
        \hskip.5\tabcolsep}
\newcommand{\cdashlinelr}[1]{%
  \noalign{\vskip\aboverulesep
           \global\let\@dashdrawstore\adl@draw
           \global\let\adl@draw\adl@drawiv}
  \cdashline{#1}
  \noalign{\global\let\adl@draw\@dashdrawstore
           \vskip\belowrulesep}}
\makeatother

\makeatletter
\newcommand{\ssymbol}[1]{^{\@fnsymbol{#1}}}

\makeatother

%% file: latex/sections/0_abstract.tex
\begin{abstract}
The choice of tokenizer can profoundly impact language model performance, 
yet accessible and reliable evaluations of tokenizer quality remain an open challenge.
Inspired by scaling consistency, we show that smaller models can accurately predict significant differences in tokenizer impact on larger models at a fraction of the compute cost.
By systematically evaluating both English-centric and multilingual tokenizers, we find that tokenizer choice has negligible effects on tasks in English but results in consistent performance differences in multilingual settings.
We propose new intrinsic tokenizer metrics inspired by Zipf's law that correlate more strongly with downstream performance than text compression when modeling unseen languages.
By combining several metrics to capture multiple aspects of tokenizer behavior, we develop a reliable framework for intrinsic tokenizer evaluations.
Our work offers a more efficient path to informed tokenizer selection in future language model development.\looseness=-1
\end{abstract}

%% file: latex/sections/1_intro.tex
\section{Introduction}
Language models rely on tokenizers to convert text into machine-interpretable tokens \citep{Grefenstette1999}. 
As tokenizers determine how text is segmented, 
typically into subword units \citep{sennrich-etal-2016-neural, kudo-2018-subword},
they fundamentally shape the statistical patterns that language models learn to estimate, thereby impacting both efficiency and downstream  performance \citep{domingo2019much,bostrom-durrett-2020-byte, ali-etal-2024-tokenizer}.
Since updating the tokenizer after model training is more cumbersome than ablating other architectural or training decisions \citep{yong-etal-2023-bloom,zhao2024llamaenglishempiricalstudy}, 
understanding tokenizer impact on model performance prior to large-scale training is crucial.\looseness=-1

Tokenizer design and evaluation remain open challenges in NLP 
\citep{gowda-may-2020-finding, cognetta-etal-2024-two}.
\textit{Extrinsic} tokenizer assessments, which involve training a model to measure the impact on performance, are prohibitively expensive for rapid iteration.
As a result, \emph{intrinsic} indicators are commonly utilized, 
with text compression often presented as a strong predictor of performance \citep{galle-2019-investigating,klein-tsarfaty-2020-getting,rust-etal-2021-good}.
However, recent studies question the robustness of only considering text compression \citep{zouhar-etal-2023-tokenization,ali-etal-2024-tokenizer,schmidt-etal-2024-tokenization}, motivating the search for more reliable frameworks.\looseness=-1

In this work, we address the practical question of how to select a tokenizer for training a decoder-only language model.
We focus on how significant differences in tokenizer quality manifest across both smaller and larger models,
informed by evidence that 
variations in design choices can be traced across model scales
\citep{zohar2024apolloexplorationvideounderstanding,choshen2024hitchhikersguidescalinglaw}.
Concretely, we examine whether performance patterns observed in 350M-parameter models, varying only in their choice of tokenizer, can predict those at the 2.7B-parameter scale. 
This approach reduces the computational cost of extrinsic evaluation by 85\% while isolating tokenizer impact, allowing for a methodical analysis of the relationship between intrinsic tokenizer characteristics and downstream performance.\looseness=-1

Recent work has largely confined tokenizer evaluations to monolingual settings
\citep{goldman-etal-2024-unpacking,schmidt-etal-2024-tokenization} or limited multilingual comparisons to classification tasks \citep{ali-etal-2024-tokenizer}.
However, as large language models (LLMs) emerge as universal task solvers \citep{openaigpt4},
we argue that tokenizer evaluation should also extend to a broader range of applications
\citep{dagan-2024-getting}.
To understand \textit{when} tokenizer choice matters, 
we first pretrain 350M-parameter and 2.7B-parameter models on English-centric data.
We then systematically evaluate tokenizer impact across four English-centric and two multilingual tokenizers on multiple-choice benchmarks, summarization, 
and machine translation into and out of
multiple languages and scripts.
Our experiments show that tokenizer choice does not have a scale-consistent 
impact on English-language tasks (likely due to sufficient vocabulary coverage), 
yet produces persistent differences in translation scenarios.
We find that a 350M-parameter model with a multilingual tokenizer can outperform a 2.7B-parameter model that uses an English-centric tokenizer, demonstrating that careful tokenizer selection can offset substantial increases in model size.\looseness=-1

To address the need for more reliable intrinsic evaluations,
we hypothesize that tokenizers yielding token distributions closely aligned with the statistical patterns of natural language are especially well-suited for generative tasks in that language.
Accordingly, we propose four new metrics based on different properties of the token distributions produced on
 downstream tasks.
Although these metrics do not significantly correlate with generative performance in English,
they offer more reliable predictions of performance than text compression when modeling previously unseen languages.
Finally, we propose a two-stage predictive framework for intrinsic evaluation, combining several tokenizer metrics to produce consistent tokenizer rankings.
Our findings offer a practical path toward better-informed tokenizer selection in future language model development.\looseness=-1

%% file: latex/sections/2_tokenizer.tex
\section{Tokenizer Choice}
\label{sec:tokenizers}
The choice of tokenizer is a fundamental decision when developing a new language model.
Although several aspects of tokenizer creation remain active areas of research \cite{schmidt-etal-2024-tokenization, ali-etal-2024-tokenizer}, 
there is broad consensus that a language model and its tokenizer should ideally be trained on the same data distribution
\citep{workshop2023bloom176bparameteropenaccessmultilingual, hayase2024data}.
Simply adopting an existing tokenizer can jeopardize this alignment, 
potentially leading to
sub-optimal text encoding \citep{ahia-etal-2023-languages} and, in severe cases,
degraded performance and unintended model behavior \citep{land2024fishingmagikarpautomaticallydetecting, geiping2024coercingllmsrevealalmost}.
At the same time, developing a custom tokenizer without careful design and validation can
introduce severe inequalities and limitations across languages \citep{ahia-etal-2023-languages,petrov2023language}.

In practice, the resource cost of designing and thoroughly evaluating a new tokenizer likely explains why some models simply borrow a pretrained one from an existing language model.
Unfortunately, the exact rationale behind such decisions is rarely documented,
leaving the impression that a choice was made after little or no systematic testing. 
This lack of transparency underscores the need for more reliable yet low-cost methods to guide tokenizer selection.\looseness=-1

For our experiments, we evaluate the tokenizers from the following published language models:
\paragraph{\llama}
English-centric; used by 
Llama and 
Llama 2 \citep{touvron2023llamaopenefficientfoundation, touvron2023llama2openfoundation}, 
ALMA \citep{xu2024a}, 
Mistral \citep{jiang2023mistral7b}, 
OpenELM \citep{mehta2024openelmefficientlanguagemodel}, and
Phi-3-mini \citep{abdin2024phi3technicalreporthighly}.\looseness=-1\footnote{For our experiments, we rely on the MIT-licensed implementation provided with the Phi-3-mini model.}
\paragraph{\textsc{GPT-2}}
English-centric; used by 
GPT-2 \citep{radford2019language},
GPT-3 \citep{NEURIPS2020_GPT3},
Megatron \citep{shoeybi2020megatronlmtrainingmultibillionparameter},
OPT \citep{zhang2022optopenpretrainedtransformer}, and
RoBERTa \citep{liu2019robertarobustlyoptimizedbert}.
\paragraph{\textsc{GPT-NeoX}} 
English-centric; used by 
GPT-NeoX \citep{black-etal-2022-gpt}, 
DCLM \citep{li2024DCLM},
OLMo \citep{groeneveld-etal-2024-olmo}, and 
Pythia \citep{biderman2023pythia}.
\paragraph{\textsc{Falcon}} 
English-centric; used by the Falcon models 
\citep{almazrouei2023falconseriesopenlanguage}.
\paragraph{\phthreesmall}
Multilingual; used by GPT-4 \citep{openaigpt4} and the basis for Llama 3 \citep{dubey2024llama3herdmodels}.\looseness=-1
\paragraph{\textsc{Aya 23}}
Multilingual coverage spanning 23 major Asian, European, and Middle Eastern languages; used by Aya 23 \citep{aryabumi2024aya}.

%% file: latex/sections/3_approach.tex
\section{Proposed Approach}
\label{sec:approach}
Our experimental approach involves pretraining language models with the chosen tokenizers at two scales, assessing their downstream performance across multiple tasks, and comparing these results with intrinsic tokenizer metrics.

\subsection{Model Architecture and Scales}
\label{subsec:arch}
NLP has largely shifted toward decoder-only architectures, driven by their emergent applicability to diverse tasks
\citep{wei2022emergent,Palm-JMLR:v24:22-1144,tay2023ul}.
Notably, recent advances in machine translation \citep{xu2024a, xu2024contrastive} challenge the traditional dominance of encoder-decoder systems \citep{bojar-etal-2018-findings, barrault-etal-2019-findings, barrault-etal-2020-findings, akhbardeh-etal-2021-findings, kocmi-etal-2022-findings}. 
Following this trend, we restrict our focus to decoder-only transformer models \citep{DBLP:conf/nips/VaswaniSPUJGKP17}.\looseness=-1

We conjecture that if a tokenizer significantly affects model quality, 
its impact will manifest consistently across different model scales.
Although differences only revealed at specific scales may be relevant \citep{tao2024scalinglawsvocabularylarger}, 
we are primarily concerned with identifying consistent patterns since the same tokenizer is often employed for models of various sizes \citep{NEURIPS2020_GPT3,touvron2023llama2openfoundation,dubey2024llama3herdmodels}.
Smaller models, with limited representational capacity \citep{kaplan2020scalinglawsneurallanguage}, are less able to compensate for sub-optimal tokenization  \citep{chai-etal-2024-tokenization},
making them particularly effective at revealing differences in tokenizer quality.
This emphasis on scaling consistency is key to efficient model development
\citep{choshen2024hitchhikersguidescalinglaw}.\looseness=-1

We consider two architecture configurations: 
a 350M-parameter model aligned with GPT-3 Medium  
and a model following the GPT-3 configuration for 2.7B trainable parameters. 
The exact parameter count, shown in \autoref{tab:number_of_trainable_parameters}, varies with vocabulary size. 
Our choice of 350M parameters is inspired by \citet{li2024DCLM}, who show that 400M-parameter models can effectively forecast performance trends in larger architectures. 
Meanwhile, 2.7B-parameter models align with recent studies demonstrating the practical efficiency of the 3B-parameter range \citep{gunasekar2023textbooksneed, li2023textbooksneediiphi15, abdin2024phi3technicalreporthighly, mehta2024openelmefficientlanguagemodel}.\looseness=-1

\begin{table}[t]
\centering
\resizebox{\columnwidth}{!}{%
\begin{tabular}{lccccc} 
\toprule
    & \multicolumn{1}{c}{\multirow{2}{*}{\begin{tabular}[c]{@{}c@{}}Vocab\\  size\end{tabular}}} & \multicolumn{2}{c}{350M} & \multicolumn{2}{c}{2.7B} \\ \cmidrule(lr){3-4} \cmidrule(lr){5-6} 
         & \multicolumn{1}{c}{} & $\vert \theta \vert$ & Hours &  $\vert \theta \vert$ & Hours \\ \midrule
\llama & \phantom032k    & 337M    &  220      & 2.6B     & 1840      \\ 
\gpt & \phantom050k    & 356M       &  210   & 2.7B       & 1650    \\
\gptneo & \phantom050k  & 356M      &  210    & 2.7B      & 1650     \\
\falcon & \phantom065k   & 371M     &  210     & 2.7B     & 1670      \\
\phthreesmall & 100k & 407M & 220 & 2.8B  & 2050  \\
\aya & 256k   & 566M      &   490   & 3.2B   &  2120       \\
\bottomrule
\end{tabular}
}
\caption{Vocabulary size, number of trainable parameters $(\vert \theta \vert)$, and cost of pretraining measured in H100 GPU hours. For simplicity, we refer to the smaller and larger model scales as \textit{"350M"} and \textit{"2.7B"} respectively.  }
\label{tab:number_of_trainable_parameters}
\end{table}

\subsection{Pretraining}
For each tokenizer, we pretrain models at the two scales. 
Our pretraining methodology follows the GPT-3 configurations for next-token prediction, with modifications for improved stability and performance.
For 350M-parameter models, we increase the maximum learning rate from \num{3e-4} to \num{9e-4} based on pilot results showing improved downstream performance. 
We adopt the weight initialization scheme of \citet{le-scao-etal-2022-language}, whose setup closely resembles ours.
Unlike the original GPT-3 architecture, 
which employs alternating dense and locally banded sparse attention \citep{child2019generatinglongsequencessparse}, 
all models utilize full attention in every layer.
All models are trained on the 100B GPT-2 tokens subset of the English-centric FineWeb dataset \citep{penedo2024finewebdatasetsdecantingweb} with a fixed batch size of 2M tokens, exceeding the $20 \times$ parameter-count guideline of \citet{NEURIPS2022_c1e2faff} and \citet{besiroglu2024chinchillascalingreplicationattempt}.
To improve training stability, beyond the increased batch size, we add an auxiliary z-loss to constrain logit magnitudes \citep{Palm-JMLR:v24:22-1144}.\footnote{We explored scaling the embeddings by $\sqrt{d}$ as suggested by \citet{takase2024spikemorestabilizingpretraining}, but found this approach significantly degraded downstream performance on generative tasks.}
As reported in \autoref{tab:number_of_trainable_parameters}, most 350M-parameter models require around 85\% fewer H100 GPU hours to train than their 2.7B-parameter counterpart.
\autoref{tab:flops} (\autoref{app:details}) presents a complementary analysis based on FLOPs, leading to the same insights. 
\looseness=-1

\subsection{Downstream Tasks}
\label{sec:finetuning}
We evaluate our models on three task categories: multiple-choice benchmarks, summarization, and machine translation. 
The emphasis on generative tasks is motivated by their demonstrated sensitivity to tokenizer quality \citep{goldman-etal-2024-unpacking}.
\begin{figure*}[t]
\centering
    \begin{subfigure}{.49\linewidth}
        \centering
        \includegraphics[width=1\textwidth]{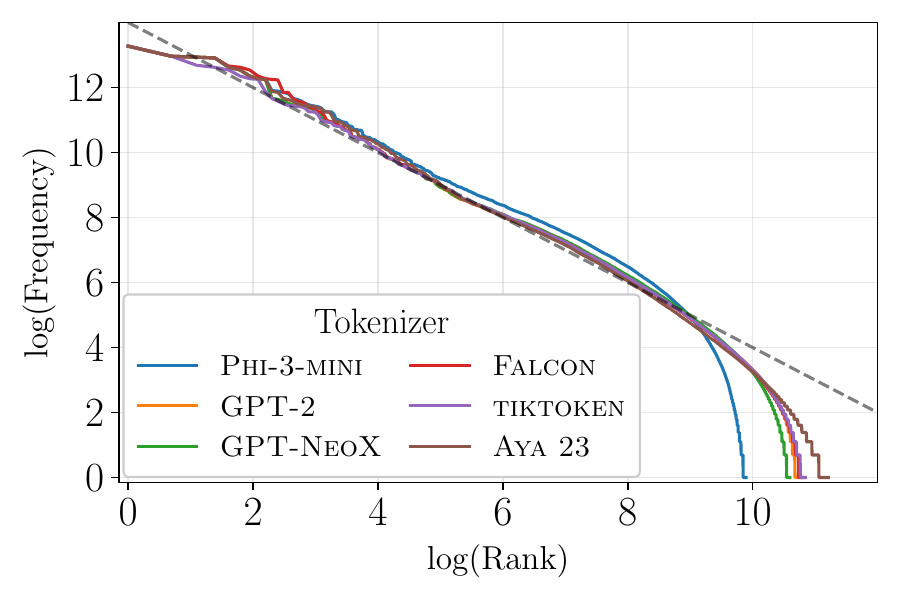} 
        \caption{\textsc{X-Sum}}
        \label{fig:xsum}
    \end{subfigure}
    \hfill 
    \begin{subfigure}{.49\linewidth}
        \centering
        \includegraphics[width=1\columnwidth]{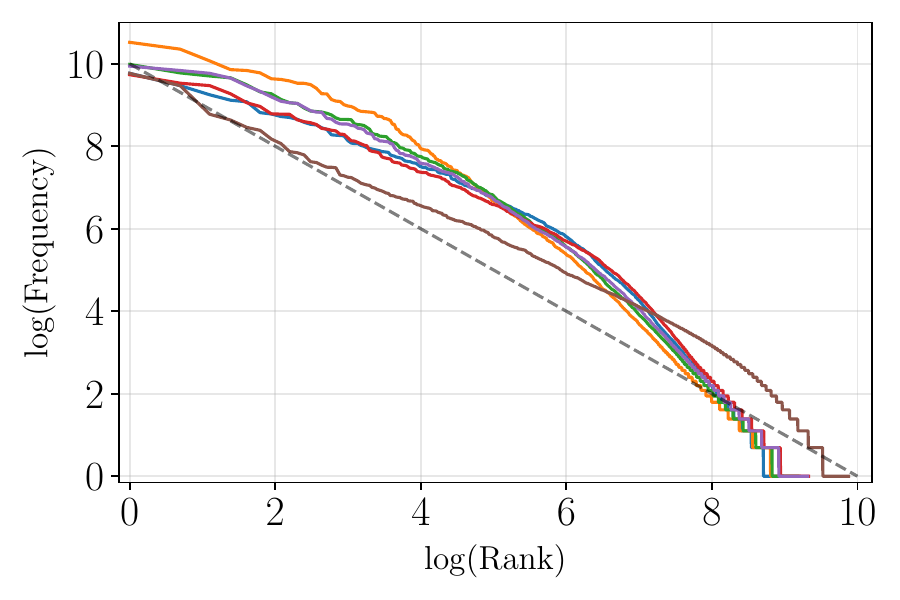} 
        \caption{\textsc{cs}}
        \label{fig:de}     
    \end{subfigure}
    \begin{subfigure}{.49\linewidth}
        \centering
        \includegraphics[width=1\columnwidth]{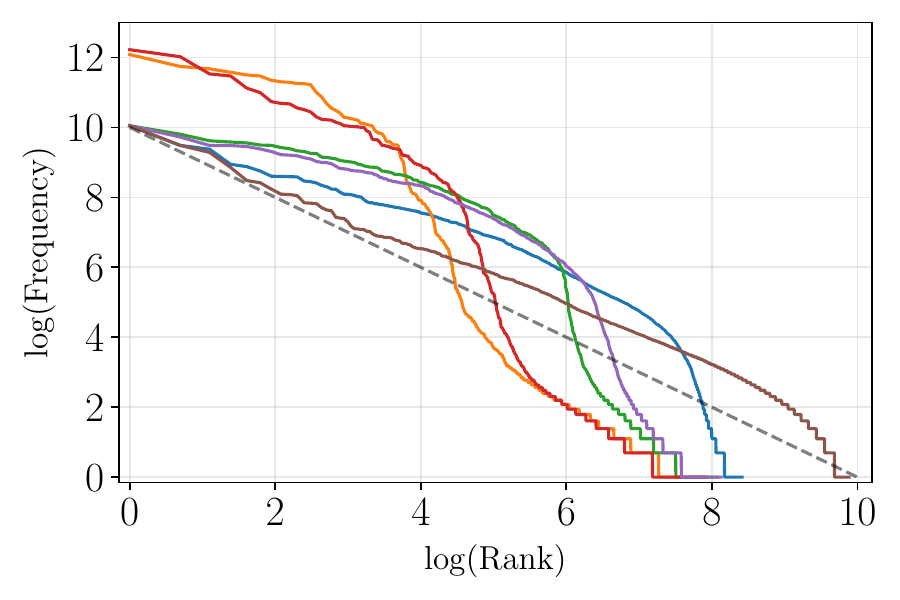} 
        \caption{\textsc{ru}}
        \label{fig:ru}     
    \end{subfigure}
    \hfill 
    \begin{subfigure}{.49\linewidth}
        \centering
        \includegraphics[width=1\columnwidth]{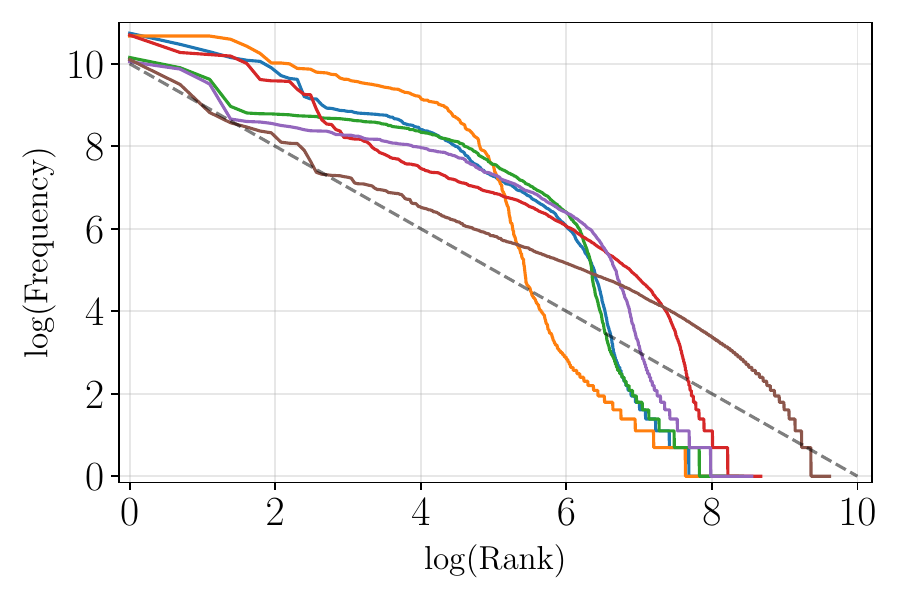} 
        \caption{\textsc{zh}}
        \label{fig:zh}     
    \end{subfigure}
    \caption{Token frequency plotted against frequency rank in log-log scale for English (\xsum),  Czech (\lcs), Russian (\lru), and Chinese (\lzh).
    The dashed lines with a slope of $-1$ reference a Zipfian power-law distribution.
    }
    \label{fig:token-distribution}
\end{figure*}

\paragraph{Multiple-choice Benchmarks}
Multiple-choice tasks represent a cornerstone of modern language model evaluations, 
serving as the primary framework for assessing zero- and few-shot capabilities across 
LLMs.
We take inspiration from the Open LLM Leaderboard (\texttt{v1})\footnote{This version of the Open LLM Leaderboard was archived in June 2024. Relevant information can be found at \url{https://huggingface.co/docs/leaderboards/en/open_llm_leaderboard/archive}.} and LLM360 \citep{liu2024llm360} 
and rely on the LM Evaluation Harness \citep{eval-harness}\footnote{Commit \texttt{dc90fec} at \url{https://github.com/EleutherAI/lm-evaluation-harness}.}
to measure performance on:\looseness=-1
\begin{itemize}[nosep,leftmargin=1em]
    \item Reasoning (R): \textsc{ARC} \citep{clark2018arc}, \textsc{HellaSwag} \citep{zellers-etal-2019-hellaswag}, \textsc{PIQA} \citep{bisk2020piqa}, and \textsc{WinoGrande} \citep{sakaguchi2021winogrande}.
    \item Knowledge understanding (KU): \textsc{MMLU} \citep{hendryckstest2021mmlu}, and \textsc{RACE} \citep{lai-etal-2017-race}.
    \item Misinformation and bias (M\&B): \textsc{CrowS-Pairs} \citep{nangia-etal-2020-crows}, and \textsc{TruthfulQA} \citep{lin-etal-2022-truthfulqa}.
\end{itemize}
We rely on the same experimental settings as used by \citet{mehta2024openelmefficientlanguagemodel}.\looseness=-1

\paragraph{Summarization}
We finetune our models on the \xsum dataset \citep{narayan-etal-2018-dont},
which tasks models with generating single-sentence summaries of news articles in English.
We evaluate the summaries with SL-BLEURT \citep{amplayo2023smart, sellam-etal-2020-bleurt}, 
which offers robust measures of coherence, factuality, fluency, and informativeness.
See \autoref{tab:xsum-details} (\autoref{app:details}) for details.\looseness=-1

\paragraph{Machine Translation}
To evaluate a tokenizer's contribution to downstream performance when modeling unseen languages and scripts,
we perform bi-directional translation experiments across four language pairs with English:
Czech (\lcs$\leftrightarrow$\len), 
German (\lde$\leftrightarrow$\len), 
Russian (\lru$\leftrightarrow$\len), 
and Chinese (\lzh$\leftrightarrow$\len). 
We finetune our models on the translation data from \citet{xu2024a} and evaluate performance on the \textsc{WMT21} test set.
Translation quality is assessed using two complementary metrics: the model-based MetricX \citep{juraska-etal-2023-metricx} and the string-based chrF \cite{popovic-2015-chrf}.
See \autoref{tab:mt-details} (\autoref{app:details}) for details.

\begin{table*}[t!]
\centering
\resizebox{\textwidth}{!}{%
\begin{tabular}{lcccccccccccccc} \toprule
 &
  \multicolumn{1}{c}{\multirow{2}{*}{\begin{tabular}[c]{@{}c@{}}Model\\ size\end{tabular}}} &
  \multicolumn{6}{c}{0-shot} & 
  \multicolumn{1}{c}{5-shot} &
  \multicolumn{1}{c}{25-shot} &   \multicolumn{1}{c}{\multirow{2}{*}{\begin{tabular}[c]{@{}c@{}}Avg.\\ \textsc{R}\end{tabular}}} &   \multicolumn{1}{c}{\multirow{2}{*}{\begin{tabular}[c]{@{}c@{}}Avg.\\ \textsc{KU}\end{tabular}}} &   \multicolumn{1}{c}{\multirow{2}{*}{\begin{tabular}[c]{@{}c@{}}Avg.\\ \textsc{M\&B}\end{tabular}}} \\ %
  \cmidrule(lr){3-8} \cmidrule(lr){9-9} \cmidrule(lr){10-10} 
 &
  \multicolumn{1}{c}{} &
  \multicolumn{1}{c}{\textsc{ARC}} &
  \multicolumn{1}{c}{\textsc{HellaSwag}} &
  \multicolumn{1}{c}{\textsc{PIQA}} &
  \multicolumn{1}{c}{\textsc{WinoGrande}} &
  \multicolumn{1}{c}{\textsc{RACE}} &
  \multicolumn{1}{c}{\textsc{TruthfulQA}} &
  \multicolumn{1}{c}{\textsc{MMLU}} &
  \multicolumn{1}{c}{\textsc{CrowS-Pairs}} %
   \\ \midrule
\llama    & 350M & 25.3 & 44.9 & 69.8 & 51.9 & 31.1 & 38.5 & 25.9 &  \textbf{63.9}  &  48.0  &  28.5  &  51.2  \\ 
\gpt   & 350M & 25.3 & 45.9 & \textbf{71.2} & \textbf{53.7} & 31.4 & \textbf{40.7} & 26.2 & 63.1  &  49.0  &  28.8  &  \textbf{51.9}  \\ 
\gptneo & 350M & 25.4 & 45.9 & 70.5 & 53.2 & \textbf{32.8} & 37.4 & \textbf{26.7} & 62.4  &  48.8  &  \textbf{29.8}  &  49.9   \\ 
\falcon   & 350M & 24.7 & 46.7 & 70.1 & 53.4 & 31.9 & 39.1 & 25.8 &  62.6 &  48.7 &  28.9  &  50.9  \\
\phthreesmall & 350M & 25.2 & 45.8 &  71.2 &  53.4 &  31.8 &  39.5 &  24.7 &  59.3 & 48.9 & 28.3 & 49.4 \\
\aya   & 350M & \textbf{26.7} & \textbf{46.9} & 70.7 & 52.3 & 32.2 & 40.2 & 25.1 & 63.0  &  \textbf{49.2}  &  28.7  &  51.6  \\
\midrule
\llama    & 2.7B & 28.7 & 60.0 & 74.2 & 56.0 & 34.6 & 36.0 & 25.4 & 67.1  &  54.7  &  30.0  &  51.6  \\ 
\gpt   & 2.7B & 28.9 & 60.2 & 75.1 & 56.8 & 35.2 & 34.1 & 26.8 & 67.1  &  55.3  &  31.0  &  50.6  \\ 
\gptneo & 2.7B & 28.2 & 60.4 & 75.3 & 58.3 & 35.6 & 34.0 & 26.8 & 66.9  &  55.6  &  31.2  &  50.5  \\ 
\falcon  & 2.7B & 28.6 & \textbf{61.9} & \textbf{75.8} & \textbf{59.0} & \textbf{35.9} & \textbf{36.3} & \textbf{27.0} & \textbf{68.5}  &  \textbf{56.3}  &  \textbf{31.5}  &  \textbf{52.4}  \\ 
\phthreesmall & 2.7B & 30.0 &  60.6 & 75.4 & 57.5 & 36.9 & 37.3 & 26.7 & 62.5 & 55.9 & 31.2 & 49.9\\
\aya   & 2.7B & \textbf{30.4} & 61.4 & 75.6 & 56.6 & 34.4 & 35.5 & 26.0 & 68.0  &  56.0  &  30.2  & 51.8   \\

\bottomrule
\end{tabular}%
}
\caption{
Downstream results on the multiple-choice benchmarks (higher is better). 
The metrics are normalized accuracy for \textsc{ARC} (on the Challenge Set), \textsc{HellaSwag}, and \textsc{PIQA};
accuracy for \textsc{WinoGrande}, \textsc{RACE}, \textsc{TruthfulQA} (on the multi-true/MC2 task), and \textsc{MMLU}; and PCT stereotype for \textsc{CrowS-Pairs} (English version).\looseness=-1
}
\label{tab:multiple-choice}
\end{table*}

\subsection{Intrinsic Metrics}
Intrinsic evaluation offers a computationally efficient way to assess tokenizer quality without the cost of model training. 
Throughout, we measure text compression as the number of tokens after tokenizing a sequence (\textsc{compression}).
To broaden our understanding of tokenizer behavior, we examine multiple aspects of tokenization beyond compression alone.\looseness=-1

From statistical linguistics, word frequency distributions of natural language follow Zipf's law \citep{Zipf35, Zipf49}, 
a pattern that holds across languages \citep{Piantadosi2014Zipf}. 
However, it remains unclear 
how well tokenizers capture this fundamental corpus statistic \citep{gerz-etal-2018-relation},
and whether preserving the natural rank-frequency distribution in the token space benefits model learning \citep{wei-etal-2021-frequency}.
We hypothesize that tokenizers yielding token distributions closely aligned with a Zipfian power law 
may be better suited for downstream modeling of natural language.
Additionally, 
the number of unique tokens produced during tokenization can be seen as an indicator for subword coverage and reliance on fallback units, such as byte-level representations.
A higher number of unique tokens might therefore indicate a closer match between the tokenizer vocabulary and the original text distribution.\looseness=-1\footnote{Additional aspects of tokenizer quality, such as handling rare words, morphological segmentation, and pre-tokenization \citep{schmidt-etal-2024-tokenization, dagan-2024-getting}, warrant dedicated investigations in future work.}

\autoref{fig:token-distribution} illustrates token frequencies against frequency rank on log-log scales for the training split of \xsum and for \lcs, \lru, and \lzh from \citet{xu2024a}.\footnote{\textsc{de} is plotted in \autoref{fig:de2} (\autoref{app:details}).}
Rank-frequency patterns for \xsum differ minimally across tokenizers, suggesting a potentially smaller performance gap on this task. 
In contrast, some variation among tokenizers can be seen for \lcs, while the two non-Latin scripts \lru and \lzh exhibit the largest distributional differences. 
Notably, among the six tokenizers, the \aya tokenizer most closely follows a Zipfian pattern for \lzh without an overrepresentation of high-frequency tokens \citep{zouhar-etal-2023-tokenization}, indicating a potential advantage on Chinese text.\looseness=-1

Motivated by these linguistic considerations and the observed tokenizer behaviors illustrated in \autoref{fig:token-distribution}, we extend our evaluations beyond text compression with four additional intrinsic metrics:
\paragraph{Number of unique tokens (\textsc{cardinality})} The cardinality of the token set after tokenization. \looseness=-1
\paragraph{Rank-frequency AUC (\textsc{auc})} 
The area under the curve of sorted token frequency against frequency rank in log-log scale (as illustrated in \autoref{fig:token-distribution}), computed using Simpson's rule.
\paragraph{Slope of linear function (\textsc{slope})} 
The slope $\beta_1$ from estimating a linear function $f(x)=\beta_0+\beta_1x$ of token frequency as a function of frequency rank in log-log scale, approximating Zipf's law. 
\paragraph{Deviation from linear function (\textsc{power law})} 
The mean absolute error 
from the estimated linear function $f(x)$,
$\frac{1}{n}\sum_{i=1}^n 
\left\vert \beta_0 + \beta_1x_i - y_i \right\vert$.
This metric quantifies how closely the token distribution aligns with a Zipfian power law.\footnote{
For \textsc{auc}, \textsc{power law}, and \textsc{slope}, motivated by the evidence that power laws only apply above some minimum 
\citep{power_laws_pareto_zipf, empirical_power_law, large_scale_zipf}, we restrict our analysis to tokens with $\log(\text{rank}) \leq 6$.
}

%% file: latex/sections/4_results.tex
\section{Experimental Results}
\label{sec:experiments}
For each task, we evaluate whether intrinsic metrics and model performance at the 350M-parameter scale can reliably predict the relative downstream performances of 2.7B-parameter models.
As the models vary only in their choice of tokenizer, the experimental setting isolates tokenizer impact on downstream performance.
We measure the monotonic relationship between intrinsic metrics and downstream performance using Spearman's $\rho$, 
and compare rankings across scales with Kendall's $\tau$.
 
\subsection{Multiple-choice Benchmarks}
\autoref{tab:multiple-choice} reports the results for the multiple-choice tasks.
As expected, larger models outperform their smaller counterparts on reasoning tasks \citep{wei2022emergent}, 
but the gap narrows for knowledge-based tasks and becomes negligible for misinformation and bias tasks.
The reduced performance  on \textsc{TruthfulQA} for larger models is attributable to the \textit{U-shaped scaling} properties of the task as identified by \citet{wei-etal-2023-inverse}.
At 2.7B parameters, the model trained with the \falcon tokenizer generally outperforms the others;
however, no clear winner emerges at the 350M scale.\looseness=-1

\autoref{tab:intrinsic} summarizes the intrinsic and extrinsic evaluations across all tasks.
For these multiple-choice benchmarks, performance at the 2.7B scale cannot be reliably extrapolated from 350M-parameter models, and 
\textsc{compression} emerges as the most significant predictor of average downstream performance.\looseness=-1

\begin{table}[t!]
\centering
\resizebox{\columnwidth}{!}{%
\setlength\tabcolsep{3pt}
\begin{tabular}{lccc} \toprule
            & \multirow{2}{*}{\begin{tabular}[c]{@{}c@{}}Multiple-\\ choice\end{tabular}}             & \multirow{2}{*}{\begin{tabular}[c]{@{}c@{}} Summarization \end{tabular}}             & \multirow{2}{*}{\begin{tabular}[c]{@{}c@{}}Machine\\ translation\end{tabular}}                \\ 
            &  &  &  \\ \midrule
\textsc{compression} &     $\bm{-0.59}^{**}$           &        $-0.09$          &          $\hphantom-0.77^{**}$       \\ 
\textsc{cardinality}    &        $\hphantom-0.29^{*\hphantom{*}}$          &        $-0.09$          &       $\bm{-0.79}^{**}$           \\
\textsc{auc}       &       $\hphantom-0.19^{\hphantom{**}}$           &         $\hphantom-0.14$         &          $\hphantom-0.77^{**}$        \\
\textsc{power law}      &         $\hphantom-0.0\hphantom0^{\hphantom{**}}$         &         $\hphantom-0.14$         &          $\hphantom-0.78^{**}$        \\ 
\textsc{slope}     &        $\hphantom-0.0\hphantom0^{\hphantom{**}}$          &        $-0.43$          &         $-0.44^{*\hphantom*}$          \\
 \midrule
Across scales          & 0.33 & $-0.07$ & $0.87    ^{*}$ \\ 
\bottomrule
\end{tabular}%
}
\caption{Correlation analysis for all downstream tasks: Spearman's $\rho$ coefficients between intrinsic metrics and downstream performance at 
2.7B parameters (top); 
Kendall's $\tau$ coefficients comparing ranked downstream performances between the two scales (bottom). Statistical significance is denoted as: $^{\ast}p < 0.05$; $^{\ast\ast}p < 0.01$.\looseness=-1
}
\label{tab:intrinsic}
\end{table}

\subsection{Summarization}
\label{sec:summ}
Results for \textsc{X-Sum} are shown in \autoref{tab:generative-results}.
At both model scales, most tokenizers yield similar performances, except for \phthreesmall underperforming especially at 2.7B.
Moreover, the \aya tokenizer demonstrates that multilingual coverage does not hinder 
English performance compared to English-centric tokenizers.

In \autoref{tab:intrinsic}, we observe an insignificant rank correlation between the two scales, 
and none of the intrinsic metrics emerge as predictive of downstream performance.
Indeed, Zipfian patterns may be less informative in English natural language domains, where all evaluated token distributions follow similar power law trends.
This result challenges previous findings suggesting that a tokenizer's compression efficiency strongly predicts success in English generation \citep{goldman-etal-2024-unpacking}.
A plausible explanation is that once compression surpasses a certain threshold, further reductions yield diminishing returns and other factors become more decisive.\looseness=-1

\subsection{Machine Translation}
\label{sec:mt-res}
\autoref{tab:generative-results} also summarizes our machine translation results (chrF scores and detailed outcomes for all translation directions are presented in \autoref{tab:full-mt} in  \autoref{app:details}).
\aya consistently outperforms the other tokenizers at both model scales for translating into and out of English.
Furthermore, the 350M-parameter model using the \aya tokenizer performs comparably to \gptneo at 2.7B and even surpasses \gpt, which features five times more trainable parameters.
This underscores that an appropriate tokenizer can compensate for a substantially smaller parameter count.
However, in addition to incurring higher pretraining costs, a larger vocabulary also  increases inference time, as detailed in \autoref{tab:speed_benchmark} (\autoref{app:details}).\looseness=-1

In \autoref{tab:intrinsic}, we observe a significant rank correlation across scales, where only \falcon and \phthreesmall change places,
indicating scale-consistent performance trends when processing non-English data.
In contrast to the English tasks, all intrinsic metrics exhibit significant correlations with 2.7B-parameter performance,
with \textsc{cardinality} demonstrating the strongest correlation overall.\looseness=-1\footnote{Performing the same analysis based on chrF instead of MetricX yields similar results.
Since MetricX achieves a stronger correlation with human judgment \citep{freitag-etal-2023-results}, we base the remainder of our analyses on MetricX.}

\begin{table}[t]
\centering
\resizebox{\columnwidth}{!}{%
\begin{tabular}{lccccc} \toprule
 & \multirow{2}{*}{\begin{tabular}[c]{@{}c@{}}Model\\ size\end{tabular}} & \multirow{2}{*}{\begin{tabular}[c]{@{}c@{}} \textsc{X-Sum} \end{tabular}} & \multicolumn{3}{c}{\textsc{WMT21}} \\  \cmidrule(lr){4-6} 
         &      &       & \textsc{en}$\rightarrow$\textsc{xx} & \textsc{xx}$\rightarrow$\textsc{en} & Avg. \\ \midrule
\llama    & 350M & 36.0 & 11.0 & \phantom08.9  & 10.0 \\
\gpt    & 350M & 37.0 & 16.9 & 12.1 & 14.5 \\
\gptneo & 350M & 37.3 & 13.0 & \phantom09.5  & 11.3 \\
\falcon   & 350M & \textbf{37.4} & 11.9 & \phantom09.5  & 10.7 \\
\phthreesmall & 350M & 36.5 & 12.4 & 10.0  & 11.2  \\ 
\aya  & 350M & 37.3 & \phantom0\textbf{9.3}  & \phantom0\textbf{8.0}  & \phantom0\textbf{8.7}  \\

\midrule
\llama    & 2.7B & 42.3 & \phantom08.5  & \phantom05.8  & \phantom07.2  \\
\gpt    & 2.7B & \textbf{43.0} & 11.9 & \phantom07.2  & \phantom09.6  \\
\gptneo & 2.7B & 42.7 & 10.8 & \phantom06.5  & \phantom08.7  \\
\falcon   & 2.7B & 42.1 & 10.2 & \phantom06.2  & \phantom08.2  \\
\phthreesmall   & 2.7B & 39.0 & \phantom09.4  & \phantom06.2  & \phantom07.8  \\
\aya   & 2.7B & 42.8 & \phantom0\textbf{8.0}  & \phantom0\textbf{5.5}  & \phantom0\textbf{6.8}  \\
\bottomrule
\end{tabular}%
}
\caption{Downstream results on the generative tasks. 
The metrics are SL-BLEURT ($\uparrow$) for \textsc{X-Sum} and MetricX ($\downarrow$) for the \textsc{WMT21} test set.
The results for \textsc{xx}$\rightarrow$\textsc{en} and  \textsc{en}$\rightarrow$\textsc{xx}  present averages over all tasks for translating into and out of English, respectively.
}
\label{tab:generative-results}
\end{table}

%% file: latex/sections/5_predicting.tex
\begin{table*}[t]
\centering
\small
\begin{tabular}{lccccccc} \toprule
          & \llama & \gpt & \gptneo & \falcon & \phthreesmall & \aya  & Avg. \\ \midrule
\textsc{compression} &    0.56    &      0.0\phantom0    &   0.86    &    0.17   &  0.67   &  \textbf{0.92}  & 0.53 \\
\textsc{auc}       &    0.56    &      0.0\phantom0    &   0.40    &    0.31   &  0.33   &  \textbf{0.92}  & 0.42 \\
\textsc{cardinality} (\textsc{c})    &    0.36    &      \textbf{0.10}    &   0.33    &    0.73   &  0.64   &  \textbf{0.92}  & 0.51 \\
\textsc{power law}  (\textsc{p})   &    \textbf{0.76}    &      0.0\phantom0    &   \textbf{1.0}\phantom0    &    \textbf{0.78}   &  \textbf{0.80}   &  0.85  & \textbf{0.70} \\
\textsc{slope}  (\textsc{s})   &    0.71    &      0.0\phantom0    &   0.67    &    0.59   &  0.67   &  0.64  & 0.55 \\
\midrule
\textsc{c + p + s}       &    0.43    &      0.67    &   0.80    &    0.83   &  0.89   &  0.92  & 0.76 \\
\bottomrule
\end{tabular}%
\caption{
Predicting the best tokenizer in pairwise comparisons.
For each tokenizer, we report the $F_1$ score from a logistic regression model estimated on the remaining tokenizers. 
The best performing setting combines \textsc{cardinality}, \textsc{power law}, and \textsc{slope} (\textsc{c + p + s}) and is estimated using an SVM with a linear kernel.
}
\label{tab:predict}
\end{table*}

\begin{table}[t]
\centering
\resizebox{\columnwidth}{!}{%
\begin{tabular}{lcccc} \toprule
                           & \textsc{cs}       & \textsc{de}       & \textsc{ru}       & \textsc{zh}       \\ \midrule
\multirow{2}{*}{1st place} & \tc{\llama}    & \tc{\llama}    & \tc{\llama}    & \tc{\aya}   \\
                           & \aya   & \textbf{\llama}    & \textbf{\llama}   & \textbf{\aya}   \\
& & & & \\
\multirow{2}{*}{2nd place} & \tc{\aya}   & \tc{\aya}   & \tc{\aya}   & \tc{\falcon}  \\
                           & \llama   & \textbf{\aya}    & \textbf{\aya}   & \llama   \\
& & & & \\
\multirow{2}{*}{3rd place} & \tc{\falcon}   & \tc{\falcon}   & \tc{\phthreesmall}  & \tc{\llama}   \\
                           & \textbf{\falcon}   & \phthreesmall    & \textbf{\phthreesmall}   & \phthreesmall   \\
& & & & \\
\multirow{2}{*}{4th place} & \tc{\phthreesmall}    & \tc{\phthreesmall} & \tc{\gptneo}   & \tc{\phthreesmall} \\
                           & \textbf{\phthreesmall}   & \falcon    & \falcon   & \falcon   \\
& & & & \\
\multirow{2}{*}{5th place} & \tc{\gpt}    & \tc{\gptneo} & \tc{\falcon}   & \tc{\gptneo} \\
                           & \gptneo   & \textbf{\gptneo}    & \gptneo   & \textbf{\gptneo}   \\
& & & & \\
\multirow{2}{*}{6th place} & \tc{\gptneo} & \tc{\gpt}    & \tc{\gpt}    & \tc{\gpt}    \\
                           & \gpt   & \textbf{\gpt}    & \textbf{\gpt}   & \textbf{\gpt}   \\ \midrule
Kendall's $\tau$                & $0.73^*$   & $0.87^{**}$    & $0.87^{**}$   & $0.73^*$   \\ 
\bottomrule
\end{tabular}%
}
\caption{
Bradley-Terry ranking of tokenizers for each language.
For every rank, we report the ground truth ranking above (marked in \tc{grey}) and the prediction below.
Correct predictions are emphasized in \textbf{bold}.
Statistical significance is denoted as: $^{\ast}p < 0.1$; $^{\ast\ast}p < 0.05$.
}
\label{tab:bradley-terry}
\end{table}

\section{Predicting Relative Performances}
\label{sec:predict}
Our extrinsic evaluations indicate that tokenizer choice does not consistently affect English-centric tasks but plays a more decisive role in multilingual settings. 
Based on our experimental results in machine translation,
we propose a framework to identify the optimal tokenizer from intrinsic metrics.\looseness=-1

The correlation analysis in Section \ref{sec:experiments} 
suggests that our proposed intrinsic metrics can be as informative as \textsc{compression}.
However, correlations alone lack nuance; for instance, \autoref{tab:intrinsic} might favor the tokenizer that produces the largest set of unique tokens.
This criteria, taken to the extreme, would imply constructing vocabularies based on whole words rather than subwords.\looseness=-1

To address these limitations, we propose a two-stage predictive framework that first models pairwise differences  
and then aggregates these into global rankings.\looseness=-1

\subsection{Pairwise Comparisons}
\label{sec:pairwise}
We assume that more informative metrics lead to better predictive performances in pairwise comparisons.
For every pair of tokenizers $(i,j)$,
we consider the difference in a given intrinsic metric $X_i - X_j$ and define a binary outcome variable $Y_{ij}$ that equals 1 if tokenizer $i$ outperforms tokenizer $j$ and 0 otherwise.
The log-odds of this outcome is modeled via logistic regression:\looseness=-1
\begin{equation*} 
\log \left( \frac{\Pr(Y_i > Y_j)}{\Pr ( Y_i < Y_j)} \right) = \beta_0 + \beta_1(X_i - X_j) 
\end{equation*}
This formulation allows us to infer the probability that tokenizer $i$ outperforms tokenizer $j$ given their difference in tokenizer characteristic.
We iteratively leave out one tokenizer for validation and estimate the logistic model on the remaining five;
with four languages (averaging results over both translating into and out of English), this yields 40 pairwise comparisons per metric.\footnote{
For every model, we perform a cross-validated search for optimal hyperparameters.  
} 
This setup simulates the scenario of comparing a new tokenizer against established baselines.\looseness=-1

Because the same tokenizers are evaluated across multiple languages, the outcomes are correlated;
we therefore focus on predictive success rather than statistical inference.
\autoref{tab:predict} reports the $F_1$ score (the harmonic mean of precision and recall) on the held-out tokenizer for each individual metric.
\textsc{power law} proves the most informative predictor on average, a finding that
contradicts the simpler correlation patterns in \autoref{tab:intrinsic}.
Moreover, combining \textsc{cardinality}, \textsc{power law}, and \textsc{slope} in a support vector machine (SVM) estimated with a linear kernel 
improves generalization, particularly when evaluating \gpt.
This underscores the potential of more nuanced intrinsic evaluations that capture multiple aspects of tokenizer behavior. 

\subsection{Global Tokenizer Ranking}
To extend pairwise outcomes into a transitive global ranking, we adopt the Bradley-Terry (BT) model \citep{bradley1952rank}.
BT is widely used to rank agents in pairwise competitions,
including LLM leaderboards and RLHF algorithms \citep{christiano2017deep, ouyang2022training}.\footnote{\url{https://lmsys.org/blog/2023-12-07-leaderboard/}}$^,$\footnote{
The main difference between BT and Elo rating, 
which has also been utilized for ranking language models 
\citep{askell2021generallanguageassistantlaboratory, bai2022traininghelpfulharmlessassistant},
is the assumption that skill levels remain static.
}
Under BT, pairwise outcomes are aggregated into latent skill ratings $\lambda_i > 0$ and the probability that tokenizer $i$ outperforms tokenizer $j$ is modeled as 
\begin{equation*}
    \Pr(i > j) = \frac{\lambda_i}{\lambda_i + \lambda_j} \: .
\end{equation*}
We derive the BT parameters by first estimating the probability that tokenizer $i$ outperforms tokenizer $j$
using an SVM with an RBF kernel over all intrinsic metrics, applying Platt scaling \citep{Platt1999} for calibration. 
We iteratively leave out one language from the estimation for evaluation. 
\autoref{tab:bradley-terry} compares the resulting BT rankings to the ground-truth, demonstrating how intrinsic metrics can be used to effectively predict tokenizer performance across languages.
This approach is particularly appealing when extensive extrinsic evaluation would be computationally prohibitive.  
One limitation is that the framework does not accommodate ties, which may produce more granular rankings than practically meaningful.\looseness=-1

%% file: latex/sections/6_discussion.tex
\section{Discussion}
The better predictive performance of more nuanced intrinsic evaluations in \S\ref{sec:predict} emphasizes the importance of capturing multiple aspects of tokenizer behavior. 
However, in practice, it is often simpler to assess tokenizer quality based on a single metric when the optimal weighting or combination of multiple metrics is unclear.
Our results indicate that deviations from a Zipfian (power-law) distribution serve as the single most informative predictor of multilingual performance (\autoref{tab:predict}).

Text compression is a practical measure of efficiency, directly impacting generation speed and computational cost. 
Meanwhile, the deviation of a token distribution from a Zipfian power law benchmarks a tokenizer's alignment with the structure of natural language. 
Data-driven subword tokenizers whose token frequencies approximate a Zipfian distribution typically achieve efficient compression by reflecting the natural frequencies of words and phrases. 
Such tokenizers must support adequate vocabulary coverage to avoid overly relying on a small set of high-frequency subword tokens \citep{zouhar-etal-2023-tokenization} and an excessively long tail of low-frequency tokens \citep{gowda-may-2020-finding}.

As illustrated in \autoref{fig:xsum} for \xsum, all evaluated tokenizers fall along a similar distributional curve, accurately indicating minimal differences for English generation tasks. 
While our findings emphasize prioritizing an appropriate token distribution, once tokenizers surpass a certain threshold of distributional alignment---where the choice of tokenizer becomes less critical---optimizing for text compression can become a secondary focus to further improve decoding efficiency. 

Future work could explore interactions between these intrinsic metrics to provide more detailed guidance.
Moreover, our analyses could be extended to investigate when and how the relative importance of these metrics changes for specialized downstream tasks, such as code generation or biomedical text analysis, where syntactic or domain-specific properties may take precedence.

%% file: latex/sections/7_related.tex
\section{Related Work}
\label{sec:rel_work}
The most widely adopted algorithms for training a tokenizer include 
byte-pair encoding \citep{sennrich-etal-2016-neural} and unigram language modeling \citep{kudo-2018-subword}.
Recently, vocabulary-free approaches for decoder-only models have been proposed \citep{tai-etal-2024-pixar,chai-etal-2024-autoregressive} by rendering text as images \citep{salesky-etal-2021-robust,rust-etal-2023-pixel}.
However, these approaches only allow for continuous input representations and still rely on a vocabulary and softmax layer for text generation tasks.
Alternatively, byte-based tokenizers \citep{xue-etal-2022-byt5} avoid large vocabularies but produce prohibitively long sequences \citep{mielke2021wordscharactersbriefhistory}.
Larger, multilingual vocabularies, while potentially beneficial for generalization, can be slower during inference 
\citep{hofmann-etal-2022-embarrassingly,sun-etal-2023-multi,petrov2023language};
our findings highlight this trade-off as well (\autoref{tab:speed_benchmark}, \autoref{app:details}).\looseness=-1

Tokenizers are traditionally evaluated by their impact on downstream tasks \citep{provilkov-etal-2020-bpe,saleva-lignos-2023-changes,yehezkel-pinter-2023-incorporating} or by how well they meet specific design criteria \citep{klein-tsarfaty-2020-getting,hofmann-etal-2021-superbizarre,beinborn-pinter-2023-analyzing}.
For instance, \citet{schmidt-etal-2024-tokenization} focus on English multiple-choice benchmarks, 
whereas \citet{goldman-etal-2024-unpacking} include generation tasks and find text compression to be a strong predictor of performance. 
In contrast, \citet{ali-etal-2024-tokenizer} report that compression is not always reliable for multilingual tasks, 
challenging its viability as a sole merit for multilingual tokenizers \citep{stollenwerk2023trainingevaluationmultilingualtokenizer,martins2024eurollmmultilinguallanguagemodels}.
\citet{dagan-2024-getting} further discuss how to overcome potential pitfalls when applying a tokenizer to a domain for which it was not designed.

\citet{gowda-may-2020-finding} recommend ensuring that tokens in the long tail of infrequent vocabulary items from a Zipfian distribution are observed at least 100 times during training, enabling the model to effectively learn their distributional properties.
Complementary, \citet{zouhar-etal-2023-tokenization} propose to use Rényi entropy, a generalization of Shannon entropy \citep{shannon1948mathematical}, as an intrinsic metric for tokenizer evaluation, arguing that efficient tokenizers produce balanced token distributions by avoiding an overrepresentation of high-frequency tokens.
However, \citet{cognetta-etal-2024-two} present counterexamples showing that increasing Rényi effiency by eliminating high-frequency tokens and redistributing their probability mass can negatively correlate with downstream performance. 
Furthermore, \citet{dagan-2024-getting} find that,
contrary to expectations,
higher Rényi entropy correlates with lower performance in code generation.

%% file: latex/sections/8_conclusion.tex
\section{Conclusion}
We presented a cost-effective approach to tokenizer selection by training 350M-parameter decoder-only models that differ only in tokenizer choice, serving as reliable proxies for 2.7B-scale performance. 
Our experiments indicate that tokenizer choice is more critical in multilingual scenarios than in tasks limited to the pretraining language (English).\looseness=-1

We proposed new intrinsic tokenizer metrics that capture how closely token distributions align with a Zipfian power law. 
These metrics proved especially useful for determining performance on previously unseen languages.
Our results highlight the importance of distinguishing between different experimental settings when evaluating tokenizers, and
emphasized that comprehensive intrinsic evaluations should consider multiple aspects of tokenizer behavior.
Finally, we presented a reliable framework for ranking tokenizers based on their intrinsic metrics.\looseness=-1

%% file: latex/sections/9_limitations.tex
\section*{Limitations}
Our study focuses on decoder-only models up to 2.7B parameters, chosen for their practical relevance. 
Although our findings provide a strong basis for evaluating tokenizer performance at this scale, 
we have not verified whether these trends hold for larger architectures. 
Prior work \citep{tao2024scalinglawsvocabularylarger} indicates that vocabulary size may need to grow with model size, 
suggesting that conclusions could differ for models beyond the scales explored here.\looseness=-1

Furthermore, while we systematically evaluate tokenizer performance on five different languages, covering three different scripts, 
the scope of our multilingual experiments remains limited.
A wider range of languages could yield different outcomes, 
especially for scripts or morphological structures not represented in our training data.

We also note that the considered multiple-choice benchmarks are known to exhibit inherent variance 
\citep{madaan2024quantifyingvarianceevaluationbenchmarks, alzahrani-etal-2024-benchmarks}, 
which may amplify or mask performance differences between tokenizers. 
The results presented here should thus be interpreted with caution and ideally verified by training multiple models with different random seeds.\looseness=-1

Finally, we did not explore the sensitivity of our results to multiple random seeds, hyperparameter configurations during downstream tasks, or variations in the pretraining pipeline. 
Although these choices kept computational demands in check, they may limit the generality of our conclusions. 
Future work could address these gaps by investigating larger model sizes, additional languages, and more exhaustive hyperparameter searches.

%% file: latex/sections/10_acknow.tex
\section*{Acknowledgements}
We thank
Matthias Sperber,
Sarthak Garg,
Frederik Kølby Christensen, and
Desmond Elliott
for helpful discussions and feedback.
Jonas F. Lotz is funded by the ROCKWOOL Foundation (grant 1242).

%% file: latex/sections/11_appendix.tex
\section{Pretraining and Experimental Details}
\label{app:details}

\begin{table}[ht]
\centering
\resizebox{\columnwidth}{!}{%
\begin{tabular}{lccccccc} 
\toprule
    & \multicolumn{1}{c}{\multirow{2}{*}{\begin{tabular}[c]{@{}c@{}}Vocab\\  size\end{tabular}}} & \multicolumn{3}{c}{350M} & \multicolumn{3}{c}{2.7B} \\ \cmidrule(lr){3-5} \cmidrule(lr){6-8} 
         & \multicolumn{1}{c}{} & $\vert \theta \vert$ & Hours & TFLOPs &  $\vert \theta \vert$ & Hours &  TFLOPs \\ \midrule
\llama & \phantom032k    & 337M    &  220  & 5.35    & 2.6B     & 1840  & 36.06    \\ 
\gpt & \phantom050k    & 356M       &  210  & 5.58  & 2.7B       & 1650  & 36.62  \\
\gptneo & \phantom050k  & 356M      &  210 & 5.58   & 2.7B      & 1650   & 36.62  \\
\falcon & \phantom065k   & 371M     &  210 & 5.77    & 2.7B     & 1670  & 37.09    \\
\phthreesmall & 100k & 407M & 220 & 6.21 & 2.8B  & 1540 & 38.19 \\
\aya & 256k   & 566M      &   490 & 8.17  & 3.2B   &  2120    & 43.10   \\
\bottomrule
\end{tabular}
}
\caption{Vocabulary size, number of trainable parameters, and cost of pretraining measured in H100 GPU hours and TFLOPs following \citet{Palm-JMLR:v24:22-1144}.}
\label{tab:flops}
\end{table}

\begin{table}[ht]
\centering
\resizebox{\columnwidth}{!}{%
\begin{tabular}{lrr} \toprule
                           & 350M & 2.7B \\ \midrule
Optimizer                  &      \multicolumn{2}{c}{AdamW \citep{loshchilov2018decoupled}}      \\
Adam $\beta$                  &   (0.9, 0.999)   &   (0.9, 0.95)   \\
Adam $\varepsilon$               &      \multicolumn{2}{c}{\num{1e-8}}      \\
Clip gradient norm               &      \multicolumn{2}{c}{1.0}      \\
Weight decay               &     \multicolumn{2}{c}{0.1}      \\
Peak LR          &   \num{9e-4}   &   \num{1.6e-4}   \\
Minimum LR          &   \num{9e-5}   &   \num{1.6e-5}   \\
LR schedule     &      \multicolumn{2}{c}{Cosine Decay \citep{loshchilov2017sgdr}}     \\
LR warmup ratio &      \multicolumn{2}{c}{0.0}      \\
Batch size                 &      \multicolumn{2}{c}{2M tokens}      \\
Tied embeddings           &      \multicolumn{2}{c}{Yes}     \\
Precision           &      \multicolumn{2}{c}{BFloat16 \citep{bfloat16}}     \\
Z-loss coefficient         &      \multicolumn{2}{c}{\num{1e-4}}     \\
Training duration         &      \multicolumn{2}{c}{One epoch (100B \gpt tokens)}     \\ 
\bottomrule
\end{tabular}%
}
\caption{Pretraining details for both model scales. The implementation takes inspiration from \url{https://github.com/karpathy/nanoGPT}.
}
\label{tab:pt-details}
\end{table}

\begin{table}[h]
\centering
\small
\begin{tabular}{lrr} \toprule
                           & 350M & 2.7B \\ \midrule
Peak LR          &   \multicolumn{2}{c}{\num{1e-4}}  \\
Minimum LR          &   \multicolumn{2}{c}{\num{1e-5}}   \\
LR schedule     &      \multicolumn{2}{c}{Cosine Decay}     \\
LR warmup steps &      \multicolumn{2}{c}{1000}      \\
Batch size                 &      \multicolumn{2}{c}{128}      \\
Precision           &      \multicolumn{2}{c}{BFloat16}     \\
Training duration         &      \multicolumn{2}{c}{10 epochs}     \\
Source prefix         &      \multicolumn{2}{c}{"Article: $\{\texttt{source} \}$"}     \\
Target prefix         &      \multicolumn{2}{c}{"Summary: $\{\texttt{target} \}$"}     \\
\bottomrule
\end{tabular}%
\caption{Finetuning details for \textsc{X-Sum}.}
\label{tab:xsum-details}
\end{table}

\begin{table}[t]
\centering
\resizebox{\columnwidth}{!}{%
\begin{tabular}{lrr} \toprule
                           & 350M & 2.7B \\ \midrule
Peak LR          & \num{4.5e-4}  & \num{8e-5}  \\
LR schedule     &      \multicolumn{2}{c}{Inverse Square-root}     \\
Batch size                 &      \multicolumn{2}{c}{256}      \\
Precision           &      \multicolumn{2}{c}{BFloat16}     \\
Training duration         &      \multicolumn{2}{c}{3 epochs}     \\
\multirow{2}{*}{Source prefix }         &      \multicolumn{2}{c}{Translate this from \{$Lang_1$\} to \{$Lang_2$\}:} \\
              & \multicolumn{2}{c}{\{$Lang_1$\}: \{$Lang_1 sentence$\}}     \\
Target prefix         &      \multicolumn{2}{c}{\{$Lang_2$\}:}     \\
\bottomrule
\end{tabular}%
}
\caption{Finetuning details for machine translation, where \{$Lang_1$\} and \{$Lang_2$\} are the source and target language, respectively, and \{$Lang_1 sentence$\} is the source sentence.
}
\label{tab:mt-details}
\end{table}

\begin{table*}[ht]
\centering
\resizebox{\textwidth}{!}{%
\begin{tabular}{lccccccccccccccccccccccc}
\toprule
\multirow{2}{*}{} & \multirow{2}{*}{\begin{tabular}[c]{@{}c@{}}Model\\ size\end{tabular}} & \multicolumn{2}{c}{\textsc{en}$\rightarrow$\textsc{cs}} & \multicolumn{2}{c}{\textsc{cs}$\rightarrow$\textsc{en}} & \multicolumn{2}{c}{\textsc{en}$\rightarrow$\textsc{de}} & \multicolumn{2}{c}{\textsc{de}$\rightarrow$\textsc{en}} & \multicolumn{2}{c}{\textsc{en}$\rightarrow$\textsc{ru}} & \multicolumn{2}{c}{\textsc{ru}$\rightarrow$\textsc{en}} & \multicolumn{2}{c}{\textsc{en}$\rightarrow$\textsc{zh}} & \multicolumn{2}{c}{\textsc{zh}$\rightarrow$\textsc{en}} & \multicolumn{2}{c}{\textsc{en}$\rightarrow$\textsc{xx}} & \multicolumn{2}{c}{\textsc{xx}$\rightarrow$\textsc{en}} & \multicolumn{2}{c}{Avg.} \\
                           &                                                                       & chrF       & MetricX      & chrF       & MetricX      & chrF       & MetricX      & chrF       & MetricX      & chrF       & MetricX      & chrF       & MetricX      & chrF       & MetricX      & chrF       & MetricX      & chrF      & MetricX      & chrF      & MetricX      & chrF      & MetricX      \\ \midrule
\llama                      & 350M                                                                  & \textbf{34.6}       & 10.87        & 30.5       & 8.2          & 48.9       & 4.95         & 40.7       & 5.88         & \textbf{32.6}       & 14.02        & 33.9       & 8.39         & 13.2       & 14.22        & 26.1       & 13.24        & \textbf{32.4}      & 11.0         & 32.8      & 8.9          & 32.6      & 10.0         \\
\gpt                      & 350M                                                                  & 24.7       & 17.27        & 21.2       & 10.81        & 40.6       & 8.39         & 29.2       & 8.00         & 28.8       & 20.98        & 18.7       & 12.78        & 9.3        & 20.96        & 17.0       & 16.66        & 25.8      & 16.9         & 21.5      & 12.1         & 23.7      & 14.5         \\
\gptneo                   & 350M                                                                  & 29.5       & 13.03        & 26.8       & 8.91         & 46.2       & 5.79         & 36.5       & 6.37         & 28.5       & 17.56        & 28.3       & 9.1          & 11.1       & 15.45        & 23.9       & 13.79        & 28.8      & 13.0         & 28.9      & 9.5          & 28.9      & 11.3         \\
\falcon                     & 350M                                                                  & 33.1       & 11.9         & 28.4       & 8.82         & 47.8       & 5.12         & 42.1       & 5.91         & 32.0       & 18.16        & 26.0       & 10.71        & 15.8       & 12.61        & 28.2       & 12.66        & 32.2      & 11.9         & 31.1      & 9.5          & 31.7      & 10.7         \\
 \phthreesmall & 350M & 29 & 13.73 & 23.78 & 9.82 & 46 & 5.72 & 35.6 & 7.01 & 29.37 & 16.64 & 27.95 & 9.55 & 14.01 & 13.52 & 25.03 & 13.76 & 29.6 & 12.4 & 28.1 & 10.0 & 28.8 & 11.2 \\
\aya                     & 350M                                                                  & 32.5       & \textbf{10.09}       &\textbf{31.3}      & \textbf{7.59}       & \textbf{49.3}       & \textbf{4.31}        & \textbf{42.6}      & \textbf{5.37}         & 27.1       & \textbf{13.87}        &\textbf{37.5}       &\textbf{7.1}         &\textbf{17.8}      & \textbf{9.02}         & \textbf{29.8}       & \textbf{11.93}        & 31.7      & \textbf{9.3}          & \textbf{35.3}      & \textbf{8.0}          & \textbf{33.5}      & \textbf{8.7}          \\

\midrule
\llama                      & 2.7B                                                                  & \textbf{37.2}       & \textbf{8.98}         & 36.8       &\textbf{5.29}         &\textbf{52.3 }     & 3.56         & \textbf{49.4}      & \textbf{3.39}         & \textbf{37.0}       & \textbf{11.64}        & 41.8       & 5.89         & 17.4       & 9.85         & 36.1       & 8.61         & \textbf{36.0}      & 8.5          & 41.0      & 5.8          & 38.5      & 7.2          \\
\gpt                     & 2.7B                                                                  & 34.3       & 11.12        & 32.8       & 6.52         & 50.5       & 4.23         & 47.1       & 4.03         & 31.2       & 18.08        & 32.0       & 7.9          & 14.0       & 14.33        & 29.5       & 10.46        & 32.5      & 11.9         & 35.4      & 7.2          & 33.9      & 9.6          \\
\gptneo                  & 2.7B                                                                  & 30.6       & 12.28        & 35.3       & 5.91         & 49.2       & 4.32         & 47.3       & 3.84         & 31.0       & 14.92        & 37.4       & 6.53         & 15.3       & 11.67        & 32.2       & 9.75         & 31.5      & 10.8         & 38.0      & 6.5          & 34.8      & 8.7          \\
\falcon                     & 2.7B                                                                  & 34.7       & 10.04        & \textbf{37.4}       & 5.38         & 52.1       & \textbf{3.51}         & 48.5       & 3.77         & 28.7       & 17.83        & 35.5       & 7.34         & 18.2       & 9.37         & 37.3       & 8.24         & 33.4      & 10.2         & 39.7      & 6.2          & 36.5      & 8.2          \\
\phthreesmall & 2.7B & 35.14 & 9.97 & 38.02 & 5.62 & 51.69 & 3.75 & 49.49 & 3.78 & 34.34 & 13.68 & 41.29 & 6.23 & 17.42 & 10.1 & 35.26 & 9.18 & 34.6 & 9.4 & 41.0 & 6.2 & 37.8 & 7.8 \\
\aya                    & 2.7B                                                                  & 35.0       & 9.47         & 37.2       & 5.36         & 51.6       & 3.60         & 49.1       & 3.41         & 32.8       & 12.09        & 41.0       & \textbf{5.74}        & \textbf{21.2}      & \textbf{6.99}         & \textbf{40.7}       & \textbf{7.44}         & 35.2      & \textbf{8.0}          & \textbf{42.0}      & \textbf{5.5}          & \textbf{38.6}      & \textbf{6.8}          \\

\bottomrule
\end{tabular}
}
\caption{Detailed machine translation results on the WMT21 test sets measured with chrF (higher is better;  \texttt{nrefs:2|case:mixed|eff:yes|nc:6|nw:0|space:no|version:2.1.0}), and  MetricX (lower is better; \texttt{version:metricX23|referenceless:no}). 
Language codes follow ISO 639-1.
}
\label{tab:full-mt}
\end{table*}

\begin{table*}[ht]
\centering
\small
\begin{tabular}{lccccccccc} \toprule
Model & Model size & \textsc{cs}$\rightarrow$\textsc{en} & \textsc{de}$\rightarrow$\textsc{en} & \textsc{en}$\rightarrow$\textsc{cs} & \textsc{en}$\rightarrow$\textsc{de} & \textsc{en}$\rightarrow$\textsc{ru} & \textsc{en}$\rightarrow$\textsc{zh} & \textsc{ru}$\rightarrow$\textsc{en} & \textsc{zh}$\rightarrow$\textsc{en} \\
\midrule
\llama & 350M & 92.1 & 95.2 & \textbf{75.3} & \textbf{89.5} & \textbf{65.8} & 61.1 & 90.7 & 79.5 \\
\gpt & 350M & 77.6 & 78.8 & 58.2 & 63.5 & 55.8 & 56.0 & 95.7 & 73.8 \\
\gptneo & 350M & 83.6 & 91.6 & 62.7 & 74.7 & 58.3 & 59.5 & 90.4 & 75.2 \\
\falcon & 350M & 78.7 & 86.4 & 64.9 & 75.8 & 53.6 & 59.4 & 104.8 & 71.9 \\
\phthreesmall & 350M & 77.2 & 75.7 & 61.3 & 70.8 & 58.9 & 59.1 & 77.1 & 65.5  \\ 
\aya & 350M & 70.0 & 76.1 & 69.4 & 83.8 & 60.3 & \textbf{71.8} & 71.6 & 61.4 \\
\midrule
\llama & 2.7B & 97.9 & 96.9 & 65.2 & 72.2 & 57.9 & 51.0 & 96.1 & 84.8 \\
\gpt & 2.7B & \textbf{115.3} & \textbf{112.2} & 53.0 & 61.2 & 44.0 & 45.7 & \textbf{169.0} & \textbf{98.3} \\
\gptneo & 2.7B & 99.0 & 95.6 & 48.8 & 61.6 & 44.1 & 44.9 & 97.3 & 71.3 \\
\falcon & 2.7B & 100.6 & 96.2 & 57.7 & 71.9 & 43.8 & 50.5 & 154.3 & 80.6 \\
\phthreesmall & 2.7B & 91.1 & 87.4 & 51.3 & 62.3 & 43.9 & 48.5 & 97.1 & 73.1  \\ 
\aya & 2.7B & 74.8 & 80.3 & 53.5 & 64.0 & 45.9 & 63.4 & 68.2 & 60.2 \\
\bottomrule
\end{tabular}%
\caption{Inference speed (tokens per second) on the WMT21 test set with a batch size of 1 on a single H100 GPU.
}
\label{tab:speed_benchmark}
\end{table*}

\label{app:finetuning}
\begin{figure*}[t]
\centering
    \begin{subfigure}{.49\linewidth}
        \centering
        \includegraphics[width=1\columnwidth]{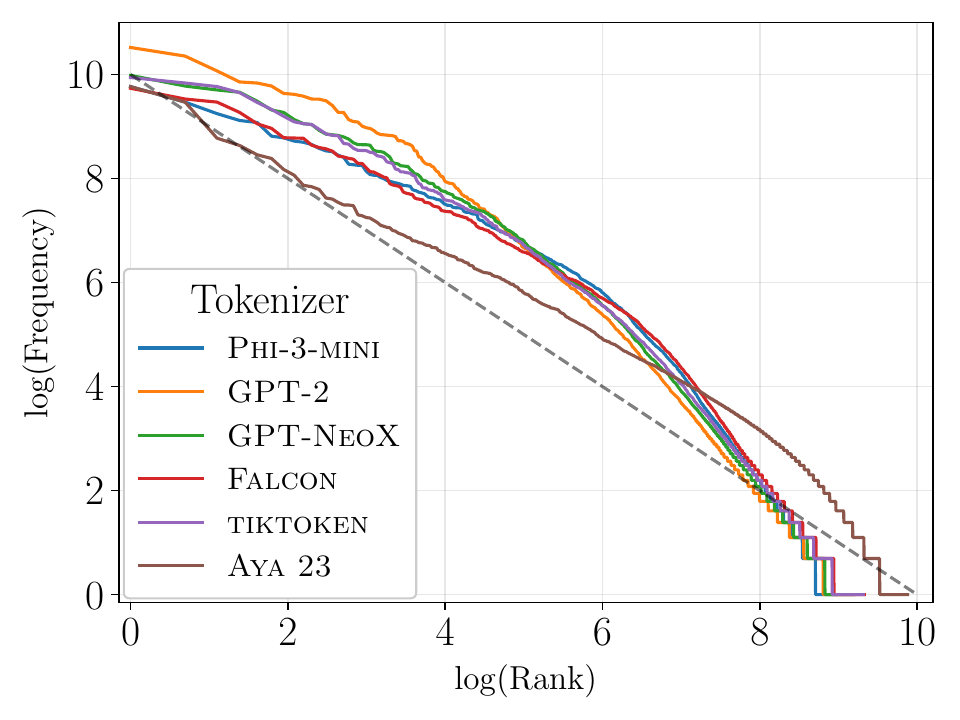} 
        \caption{\textsc{cs}}
        \label{fig:de2}     
    \end{subfigure}
    \begin{subfigure}{.49\linewidth}
        \centering
        \includegraphics[width=1\textwidth]{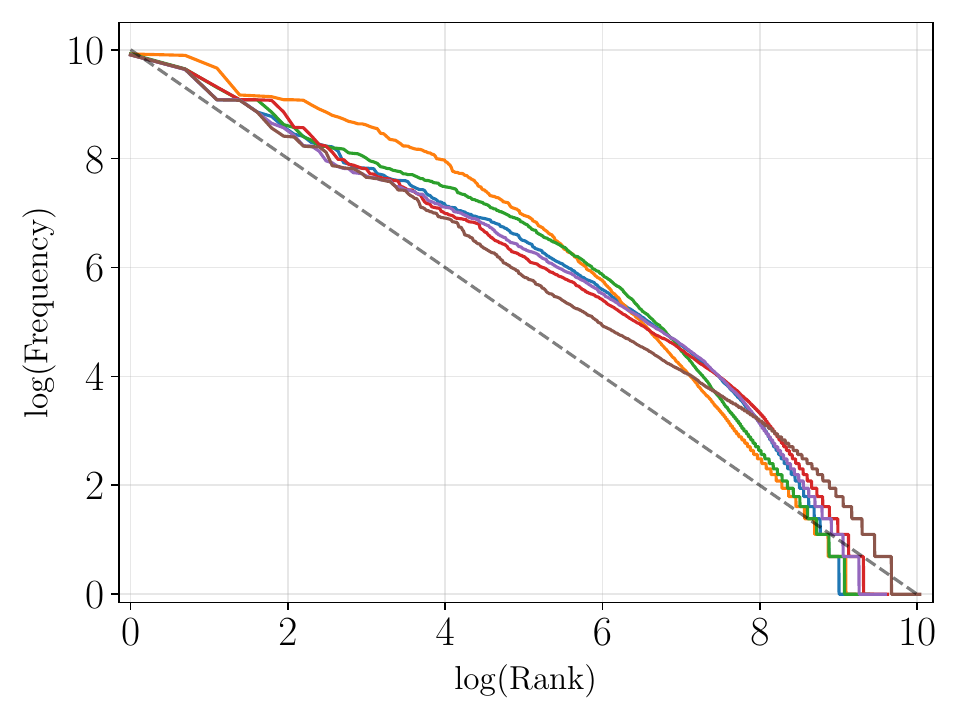} 
        \caption{\textsc{de}}
        \label{xsum2}
    \end{subfigure}
    \hfill 
    \begin{subfigure}{.49\linewidth}
        \centering
        \includegraphics[width=1\columnwidth]{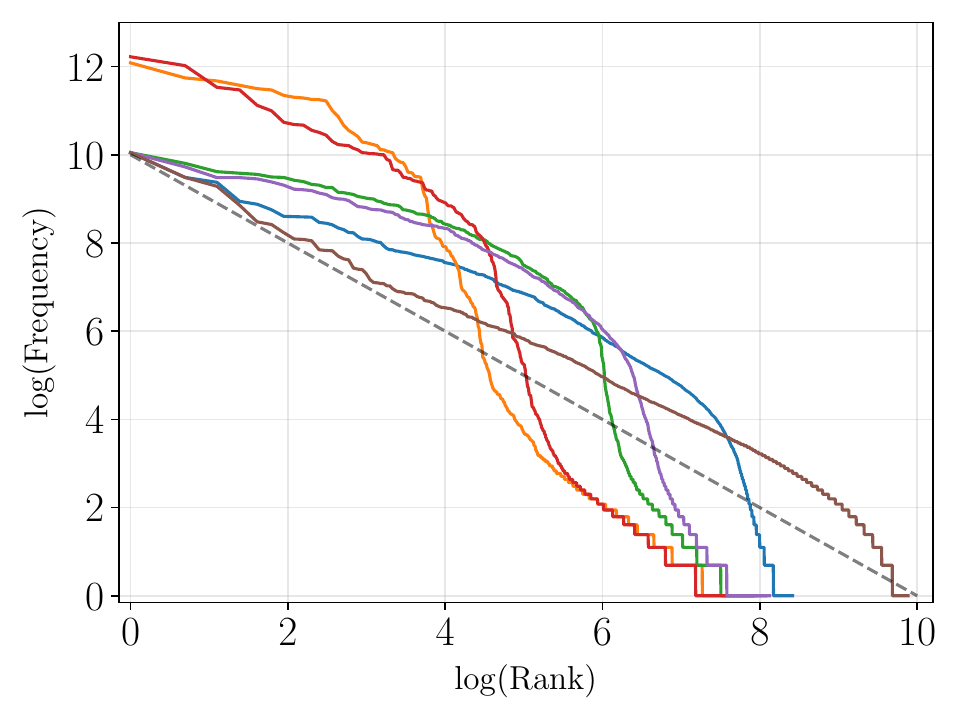} 
        \caption{\textsc{ru}}
        \label{fig:ru2}     
    \end{subfigure}
    \hfill 
    \begin{subfigure}{.49\linewidth}
        \centering
        \includegraphics[width=1\columnwidth]{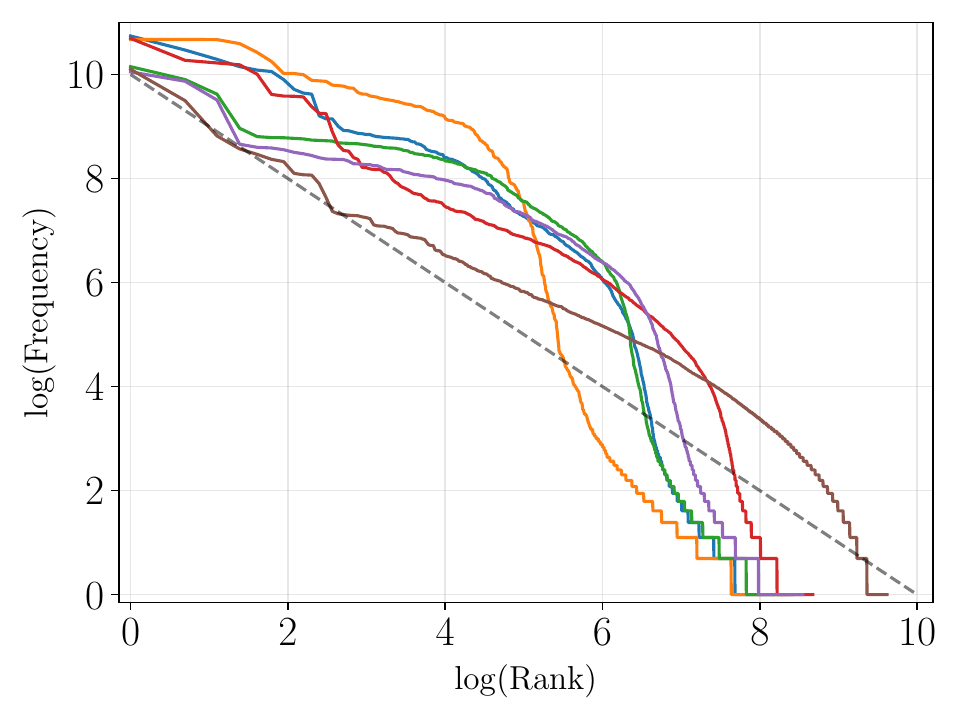} 
        \caption{\textsc{zh}}
        \label{fig:zh2}     
    \end{subfigure}
    \caption{Token frequency plotted against frequency rank in log-log scale for Czech (\lcs), German (\lde), Russian (\lru), and Chinese (\lzh).
    The dashed lines with a slope of $-1$ reference a Zipfian power-law distribution.
    }
    \label{fig:token-distribution2}
\end{figure*}